%% file: arxiv.tex
\documentclass[10pt,twocolumn,letterpaper]{article}

\usepackage[pagenumbers]{cvpr} 
\usepackage{graphicx}
\usepackage{amsmath}
\usepackage{amssymb}
\usepackage{booktabs}
\usepackage{tabu}
\usepackage{multirow}
\usepackage{float}
\usepackage{tikz}
\PassOptionsToPackage{table, dvipsnames}{xcolor}
\usepackage[table]{xcolor}
\usepackage[dvipsnames]{xcolor}
\usepackage{colortbl}
\definecolor{pink1}{rgb}{1.0, 0.51, 0.58}
\usepackage[pagebackref,breaklinks,colorlinks]{hyperref}

\usepackage[capitalize]{cleveref}
\crefname{section}{Sec.}{Secs.}
\Crefname{section}{Section}{Sections}
\Crefname{table}{Table}{Tables}
\crefname{table}{Tab.}{Tabs.}
\definecolor{lightblue}{rgb}{0.88, 0.96, 1.0}
\definecolor{lightpurple}{rgb}{0.75, 0.5, 0.75}

\begin{document}

\title{\hspace{1.7cm} Panacea: Panoramic and Controllable Video Generation for Autonomous Driving}

\author{
Yuqing Wen$^{1}$\footnotemark[1]  \footnotemark[2], Yucheng Zhao$^{2}$\footnotemark[1],  Yingfei Liu$^{2}$\footnotemark[1], Fan Jia$^{2}$, Yanhui Wang$^{1}$, Chong Luo$^{1}$\\
Chi Zhang$^{3}$, Tiancai Wang$^{2}$\footnotemark[3], Xiaoyan Sun$^{1}$\footnotemark[3], Xiangyu Zhang$^{2}$ \\
$^{1}$University of Science and Technology of China \quad  $^{2}$MEGVII Technology \quad $^{3}$Mach Drive \\
\\
\textit{Project Page}:~\href{https://panacea-ad.github.io/}{https://panacea-ad.github.io/} 
}
\makeatletter
\g@addto@macro\@maketitle{
\vspace{-1.0cm}
\begin{tikzpicture}[remember picture,overlay,shift={(current page.north west)}]
\node[anchor=north west, xshift=3.2cm, yshift=-2.8cm]{\scalebox{1}[1]{\includegraphics[width=1.85cm]{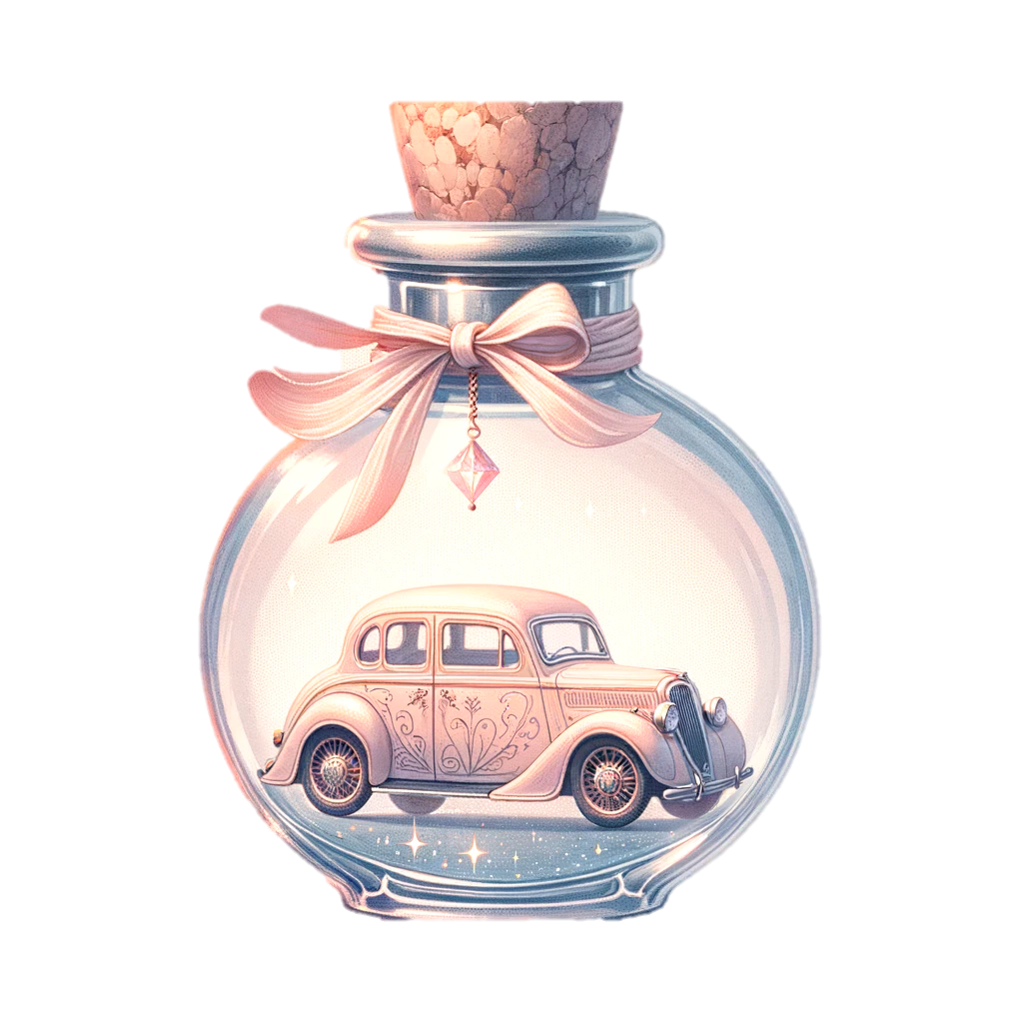}}};
\end{tikzpicture}

  \begin{figure}[H]
  \setlength{\linewidth}{\textwidth}
  \setlength{\hsize}{\textwidth}
  \centering
     \includegraphics[width=0.98\linewidth]{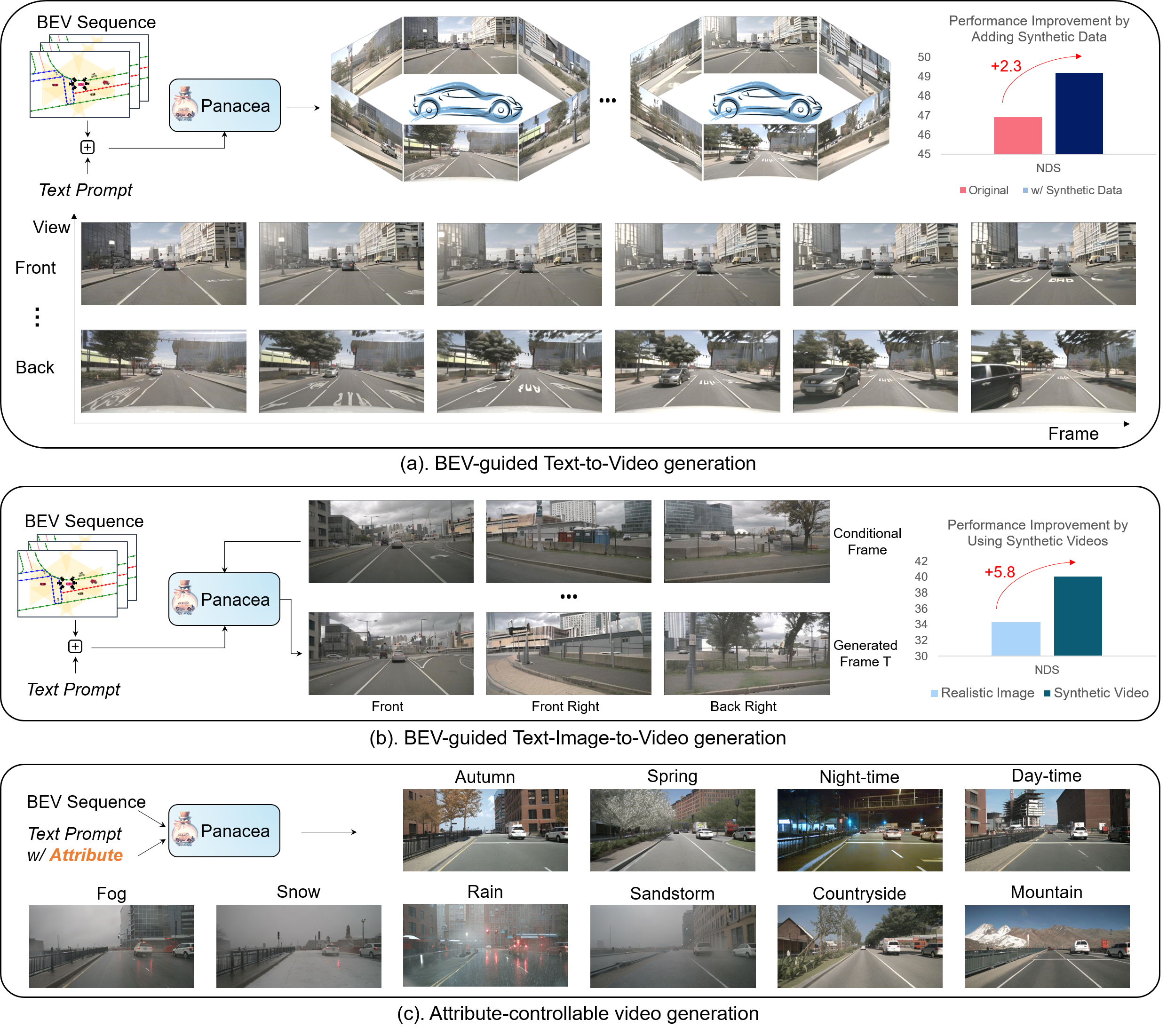}
  
  \caption{\textbf{Visualizations of the Panacea's capability.} (a). Panoramic video generation based on BEV (Bird's-Eye-View) layout sequence facilitates the establishment of a synthetic video dataset, which enhances perceptual tasks. (b). Producing panoramic videos with conditional images and BEV layouts can effectively elevate image-only datasets to video datasets, thus enabling the advancement of video-based perception techniques. (c). Video generation with variable attribute controls, such as weather, time, and scene.}

  \label{fig:fig1}
  \end{figure}
}

\makeatother
\maketitle

\renewcommand{\thefootnote}{\fnsymbol{footnote}}
\footnotetext[1]{Equal contribution.}
\footnotetext[2]{This work was done during the internship at MEGVII Technology.}
\footnotetext[3]{Corresponding author.}
\renewcommand{\thefootnote}{\arabic{footnote}}

\input{sections/0_abs}

\input{sections/1_intro}
\input{sections/2_related}

\input{sections/3_method}

\input{sections/4_experiments}
\input{sections/5_conclusion}

{\small
\bibliographystyle{ieee_fullname}
\bibliography{egbib}
}
\newpage
\appendix
\section*{Appendix}

\input{sup/sup}

\end{document}

%% file: sections/0_abs.tex
\begin{abstract}

The field of autonomous driving increasingly demands high-quality annotated training data. In this paper, we propose Panacea, an innovative approach to generate panoramic and controllable videos in driving scenarios, capable of yielding an unlimited numbers of diverse, annotated samples pivotal for autonomous driving advancements. Panacea addresses two critical challenges: 'Consistency' and 'Controllability.' Consistency ensures temporal and cross-view coherence, while Controllability ensures the alignment of generated content with corresponding annotations.  Our approach integrates a novel 4D attention and a two-stage generation pipeline to maintain coherence, supplemented by the ControlNet framework for meticulous control by the Bird's-Eye-View (BEV) layouts. Extensive qualitative and quantitative evaluations of Panacea on the nuScenes dataset prove its effectiveness in generating high-quality multi-view driving-scene videos. This work notably propels the field of autonomous driving by effectively augmenting the training dataset used for advanced BEV perception techniques.

\end{abstract}

%% file: sections/1_intro.tex
\section{Introduction}

In the domain of autonomous driving, there has been a surge of interest in Bird's-Eye-View (BEV) perception methods, which have demonstrated significant potential across key perception tasks including 3D detection~\cite{huang2021bevdet, DBLP:journals/corr/abs-2303-11926, DBLP:conf/eccv/LiWLXSLQD22}, map segmentation~\cite{liu2023petrv2, jiang2023polarformer}, and 3D lane detection~\cite{chen2022persformer, huang2023anchor3dlane}. Cutting-edge BEV perception approaches, exemplified by StreamPETR~\cite{DBLP:journals/corr/abs-2303-11926}, are trained on multi-view videos. As a result, the crux of building robust  autonomous driving system lies in high-quality, large-scale annotated video datasets. Yet, the acquisition and annotation of such data present formidable challenges. Assembling diverse video datasets encompassing a spectrum of weather, environmental, and lighting conditions not only poses challenges but can occasionally entail risks. Moreover, the annotation of video data necessitates significant resources in both effort and cost.

Inspired by the success of leveraging synthetic street images to improve the performance of perception tasks~\cite{DBLP:journals/corr/abs-2308-01661, DBLP:journals/corr/abs-2301-04634, DBLP:conf/cvpr/ZhengZLQSL23,DBLP:journals/corr/abs-2302-08908}, our proposal focuses on generating synthetic multi-view driving video data to bolster the training of cutting-edge video-based perception methods. To mitigate high annotation costs, we aim to utilize BEV layout sequences, which encompass 3D bounding boxes and road maps, for the generation of corresponding videos. Such BEV sequences can be acquired from annotated video datasets~\cite{xiao2021pandaset,sun2020scalability, DBLP:conf/cvpr/CaesarBLVLXKPBB20} or synthesized using advanced simulators~\cite{yang2023unisim,wu2023mars,heiden2021neuralsim}. This initiative can therefore be formulated as the diverse multi-view driving video generation conditioned on BEV sequences. The effectiveness of our generation model rests on two critical criteria: \textbf{controllability} and \textbf{consistency}. Empowering users to govern the generated videos via input BEV sequences and descriptive text prompts define controllability, while consistency underscores temporal coherence within individual single-view videos and coherence across multiple views.

Thanks to recent advancements in diffusion-based generative models and their extensions~\cite{DBLP:conf/iclr/SongME21,DBLP:conf/nips/HoJA20,DBLP:conf/cvpr/RombachBLEO22,DBLP:journals/corr/abs-2302-05543}, what was once a formidable challenge has become more tractable. In particular, Stable Diffusion~\cite{DBLP:conf/cvpr/RombachBLEO22} pioneers the adoption of diffusion models within latent spaces, amplifying computational efficiency with minimal compromise in generation quality. Extending this innovation, Video Latent Diffusion Model~\cite{DBLP:conf/cvpr/BlattmannRLD0FK23} expands the paradigm to high-resolution video generation by integrating temporal dimensions into established image frameworks. Furthermore, ControlNet \cite{DBLP:journals/corr/abs-2302-05543} introduces an innovative neural architecture adept at modulating pretrained diffusion models, significantly enhancing their controllability and unlocking pathways for advanced applications. Nonetheless, seamlessly amalgamating these technologies to achieve  panoramic and controllable video generation remains a huge challenge. 

In this paper, we present Panacea, an innovative video generation approach tailored specifically for panoramic and controllable driving scene synthesis. Panacea operates as a two-stage system: the initial stage crafts realistic multi-view driving scene images, while the subsequent stage expands these images along the temporal axis to create video sequences. For panoramic video generation, Panacea introduces decomposed 4D attention, enhancing both multi-view and temporal coherence. Moreover, we employ ControlNet\cite{DBLP:journals/corr/abs-2302-05543} to allow for the injection of BEV sequences. Beyond these core designs, our model retains the versatility to manipulate global scene attributes via textual descriptions, such as weather, time, and scene, offering a user-friendly interface for generating specific samples.

We apply the Panacea approach to the widely used nuScenes dateset \cite{DBLP:conf/cvpr/CaesarBLVLXKPBB20}. Comprehensive evaluations across diverse real-world application scenarios indicate Panacea's proficiency in generating valuable video training instances. Not only does it enrich existing video datasets with a plethora of synthesized samples, but it also has the potential to elevate image-only datasets to video datasets, enabling the advancement of video-based perception techniques. Furthermore, the exceptional generation fidelity coupled with heightened controllability positions Panacea as a viable candidate for real-world driving simulation. Overall, our key contributions are two-fold:

\begin{itemize}

\item We introduce Panacea, an innovative approach to multi-view video generation for driving scenes. This two-stage framework seamlessly integrates existing visual generation technologies while making important technical advancements to achieve multi-view and temporal consistency, alongside critical controllability. These technical enhancements play an important role in the solution's success.

\item Through comprehensive qualitative and quantitative assessments, Panacea demonstrates its proficiency in producing high-quality driving-scene videos. Particularly significant is the quantitative evidence highlighting the substantial enhancement our synthesized video instances provide to state-of-the-art BEV perception models. Our plan to release these synthesized instances as the "Gen-nuScenes" dataset aims to promote further research in the field of video generation for autonomous driving.
\end{itemize}

%% file: sections/2_related.tex
\section{Related Work}

\begin{figure*}[t]
  \centering
   \includegraphics[width=\linewidth]{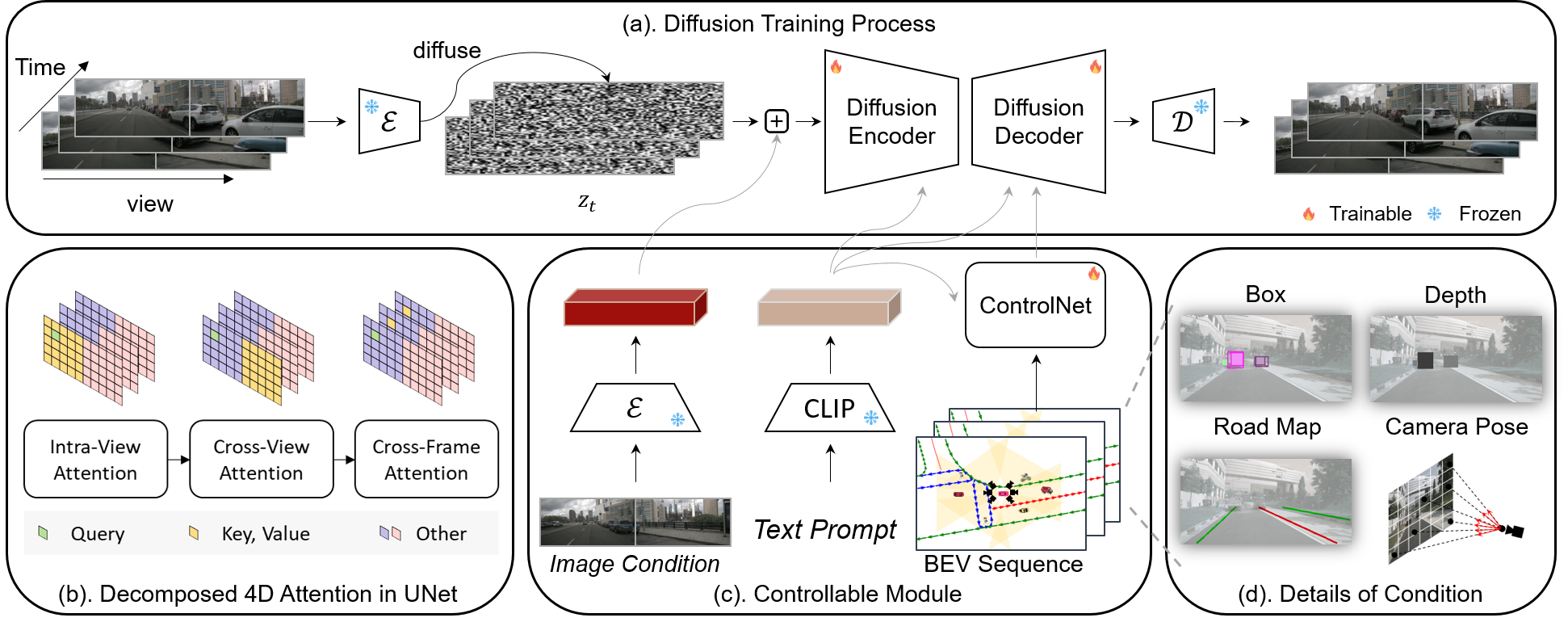}
   \hfill
    \caption{\textbf{Overview of Panacea.} (a). The diffusion training process of Panacea, enabled by a diffusion encoder and decoder with the decomposed 4D attention module. 
    (b). The decomposed 4D attention module comprises three components: intra-view attention for spatial processing within individual views, cross-view attention to engage with adjacent views, and cross-frame attention for temporal processing. 
    (c). Controllable module for the integration of diverse signals. The image conditions are derived from a frozen VAE encoder and combined with diffused noises. The text prompts are processed through a frozen CLIP encoder, while BEV sequences are handled via ControlNet. 
    (d). The details of BEV layout sequences, including projected bounding boxes, object depths, road maps and camera poses.
   }
  
   \label{fig:train}
\end{figure*}

\subsection{Diffusion-based Generative Models}
Recent advancements in diffusion models (DMs) have achieved remarkable milestones in image generation \cite{DBLP:conf/icml/NicholDRSMMSC22, DBLP:journals/corr/abs-2204-06125,DBLP:conf/nips/SahariaCSLWDGLA22,DBLP:journals/corr/abs-2307-01952, DBLP:conf/nips/DhariwalN21}. In particular, Stable Diffusion (SD) \cite{DBLP:conf/cvpr/RombachBLEO22}  employs DMs within the latent space of autoencoders, striking a balance between computational efficiency and high image quality. Beyond text conditioning, the field is evolving with the introduction of additional control signals
 \cite{DBLP:journals/corr/abs-2302-05543, DBLP:journals/corr/abs-2302-08453}.  A noteworthy example among them is ControlNet \cite{DBLP:journals/corr/abs-2302-05543}, which incorporates a trainable copy of the SD encoder to integrate control signals. Furthermore, some studies concentrate on generating multi-view images. MVDffusion \cite{DBLP:journals/corr/abs-2307-01097}, for instance, processes perspective images in parallel with a pretrained diffusion model.

In addition to image generation, the application of diffusion models for video generation \cite{DBLP:journals/corr/abs-2210-02303,DBLP:journals/corr/abs-2212-11565, DBLP:journals/corr/abs-2211-11018, DBLP:journals/corr/abs-2310-10769, DBLP:journals/corr/abs-2305-10474, DBLP:journals/corr/abs-2308-06571, DBLP:journals/corr/abs-2309-00398} is gaining significant attention. 
MagicVideo \cite{DBLP:journals/corr/abs-2211-11018} employes frame-wise adaptors and causal temporal attention module for text-to-video generation. Video Latent Diffusion Model (VLDM)~\cite{DBLP:conf/cvpr/BlattmannRLD0FK23} integrates temporal layers into a 2D diffusion model to produce temporally aligned videos.  Make-A-Video \cite{DBLP:conf/iclr/SingerPH00ZHYAG23} stretches a diffusion based text-to-image model without the necessity for text-video pairs. Imagen Video \cite{DBLP:journals/corr/abs-2210-02303} harnesses a chain of video diffusion models for generating videos according to text inputs. 

Our method also falls under the category of video generation. However, unlike previous work, we focus on the creation of controllable multi-view videos within driving contexts, representing an innovative yet complex scenario.

\subsection{Generation for Autonomous Driving}
The development of Bird's-Eye-View (BEV) representation \cite{DBLP:conf/eccv/LiWLXSLQD22, liu2022petr,DBLP:conf/aaai/LiGYYWSSL23,DBLP:journals/corr/abs-2303-11926, jiang2023far3d} in multi-view perception has become a critical research area in autonomous driving. This advancement is instrumental in enhancing downstream tasks such as multi-object tracking \cite{DBLP:conf/cvpr/ZhangCWWZ22, DBLP:conf/cvpr/PangLT0ZW23}, motion prediction \cite{DBLP:conf/cvpr/PangLT0ZW23}, and planning \cite{DBLP:conf/cvpr/HuYCLSZCDLWLJLD23,DBLP:journals/corr/abs-2303-12077}. Recently, Video-based BEV perception methods~\cite{DBLP:conf/eccv/LiWLXSLQD22,DBLP:conf/aaai/LiGYYWSSL23,DBLP:conf/iclr/ParkXYKKTZ23,DBLP:journals/corr/abs-2303-11926} have become the mainstream.  BEVFormer~\cite{DBLP:conf/eccv/LiWLXSLQD22} pioneering in integrating temporal modeling mechanism, yielding a substantial enhancement over single-frame methods such as DETR3D~\cite{DBLP:conf/corl/WangGZWZ021} and PETR~\cite{liu2022petr}. Then BEVDepth~\cite{DBLP:conf/aaai/LiGYYWSSL23}, SOLOFusion~\cite{DBLP:conf/iclr/ParkXYKKTZ23}, and StreamPETR~\cite{DBLP:journals/corr/abs-2303-11926} further improve the temporal modeling approach, achieving superior performance.

As the BEV perception methods rely heavily on paired data and BEV ground truth layouts, numerous studies are delving into paired data generation to aid training. Previously, generative efforts in autonomous driving primarily employed BEV layouts to augment image data with  synthetic single or multi-view images~\cite{DBLP:journals/corr/abs-2308-01661, DBLP:journals/corr/abs-2301-04634, DBLP:conf/cvpr/ZhengZLQSL23,DBLP:journals/corr/abs-2302-08908}, proving beneficial for single-frame perception methods. For example, BEVGen \cite{DBLP:journals/corr/abs-2301-04634} specializes in generating multi-view street images based on BEV layouts, while BEVControl \cite{DBLP:journals/corr/abs-2308-01661} proposes a two-stage generative pipeline for creating image foregrounds and backgrounds from BEV layouts. However, the generation of paired video data, crucial for more advanced video-based BEV perception methods, remains largely unexplored.
The Video Latent Diffusion Model~\cite{DBLP:conf/cvpr/BlattmannRLD0FK23} attempts to generate driving videos but its scope is limited to single-view and falls short in effectively bolstering video perception models.

In light of this, our work initiates the first exploration of generating multi-view videos paired with BEV layout sequences, marking a significant leap forward in enhancing video-based BEV perception.

%% file: sections/3_method.tex
\section{Method}

In this section, we present Panacea, an innovative approach to generate controllable multi-view videos for driving scenes. Sec. \ref{preliminary} provides a brief description of the latent diffusion models that form the foundation of our approach. Following this, Sec. \ref{framework} delves into our novel method that enables the generation of high-quality multi-view videos in a feasible and efficient manner. Finally, Sec. \ref{control} elaborates on the controlling modules integral to Panacea, which is the central design feature that renders our model an invaluable asset for the advancement of autonomous driving systems.

\subsection{Preliminary: Latent Diffusion Models} \label{preliminary}

\noindent\textbf{Diffusion models (DMs)}~\cite{DBLP:conf/nips/HoJA20, DBLP:conf/iclr/SongME21} learn to approximate a data distribution $p(x)$ via iteratively denoising a normally distributed noise $\epsilon$. Specifically, DMs first construct the diffused inputs $x_t$ through a fixed forward diffusion process in Eq.~\ref{eq:add noise}. Here $\alpha_t$ and $\sigma_t$ represent the given noise schedule, and $t$ indicates the diffusion time step. Then, a denoiser model $\epsilon_\theta$ is trained to estimate the added noise $\epsilon$ from the diffused inputs $x_t$. This is achieved by minimizing the mean-square error, as detailed in Eq.~\ref{eq:objective}. Once trained, DMs are able to synthesize a new data $x_0$ from random noise $x_T\sim \mathcal{N}(\mathbf{0}, \boldsymbol{I}) $ by sampling $x_t$ iteratively, as formulated in Eq.~\ref{eq:distribution}. Here $\mu_{\theta}$ and $\Sigma_{\theta}$ are determined through the denoiser model $\epsilon_\theta$ \cite{DBLP:conf/nips/HoJA20}.
\begin{equation}
\label{eq:add noise}
x_{t}=\alpha_{t} x+\sigma_{t} \epsilon,  \epsilon \sim \mathcal{N}(\mathbf{0}, \boldsymbol{I}), x \sim p(x)
\end{equation}
\begin{equation}
\label{eq:objective}
\min_{\theta}\mathbb{E}_{t,x,\epsilon}||\epsilon-\epsilon_{\theta}(\mathbf{x}_{t}, t)||_{2}^{2}
\end{equation}
\begin{equation}
\label{eq:distribution}
    p_{\theta}\left(x_{t-1} \mid x_{t}\right)=\mathcal{N}\left(x_{t-1} ; \mu_{\theta}\left(x_{t}, t\right), \Sigma_{\theta}\left(x_{t}, t\right)\right)
\end{equation}

\noindent\textbf{Latent diffusion models (LDMs) }~\cite{DBLP:conf/cvpr/RombachBLEO22} are a variant of diffusion models that operate within the latent representation space rather than the pixel space, effectively simplifying the challenge of handling high-dimensional data. This is achieved by transforming pixel-space image into more compact latent representations via a perceptual compression model. Specifically, for an image \( x \), this model employs an encoder \( \mathcal{E} \) to map \( x \) into the latent space \( z = \mathcal{E}(x) \). This latent code \( z \) can be subsequently reconstructed back to the original image \( x \) through a decoder \( \mathcal{D} \) as \( x = \mathcal{D}(z) \). The training and inference processes of LDMs closely mirror those of traditional DMs, as delineated in Eq. \ref{eq:add noise}-\ref{eq:distribution}, except for the substitution of \( x \) with the latent code \( z \).

\subsection{Generating High-Quality Multi-View Videos}\label{framework}

Here we describe how we upgrade a pre-trained image LDM~\cite{DBLP:conf/cvpr/RombachBLEO22} for high-quality multi-view video generation. Our model utilizes a multi-view video dataset $p_{data}$ for training. Each video sequence, encompasses \( T \) frames, indicating the sequence length, \( V \) different views, and dimensions \( H \) and \( W \) for height and width, respectively.

Our framework is built on the Stable Diffusion (SD)~\cite{DBLP:conf/cvpr/RombachBLEO22} model, which is a strong pre-trained latent diffusion model for image synthesis. While the SD model excels in image generation, its direct application falls short in producing consistent multi-view videos due to the lack of constraints between different views and frames in the sequence. Therefore, we introduce an innovative architecture: a decomposed 4D attention-based UNet, designed to concurrently generate the entire multi-view video sequence. The joint diffused input \( z \) is structured with dimensions \( H \times (W \times V) \times T \times C \), where $C$ represents the latent dimension. This multi-view video sequence is constructed by concatenating the frames across their width, which aligns with their inherent panoramic nature. Fig.~\ref{fig:train} (a) illustrates the overall training framework of the proposed model. Beyond the proposed 4D attention-based UNet, we also introduce a two-stage generation pipeline, which largely boosts the generation quality.

\begin{figure}
  \centering
   \includegraphics[width=1.0\linewidth]{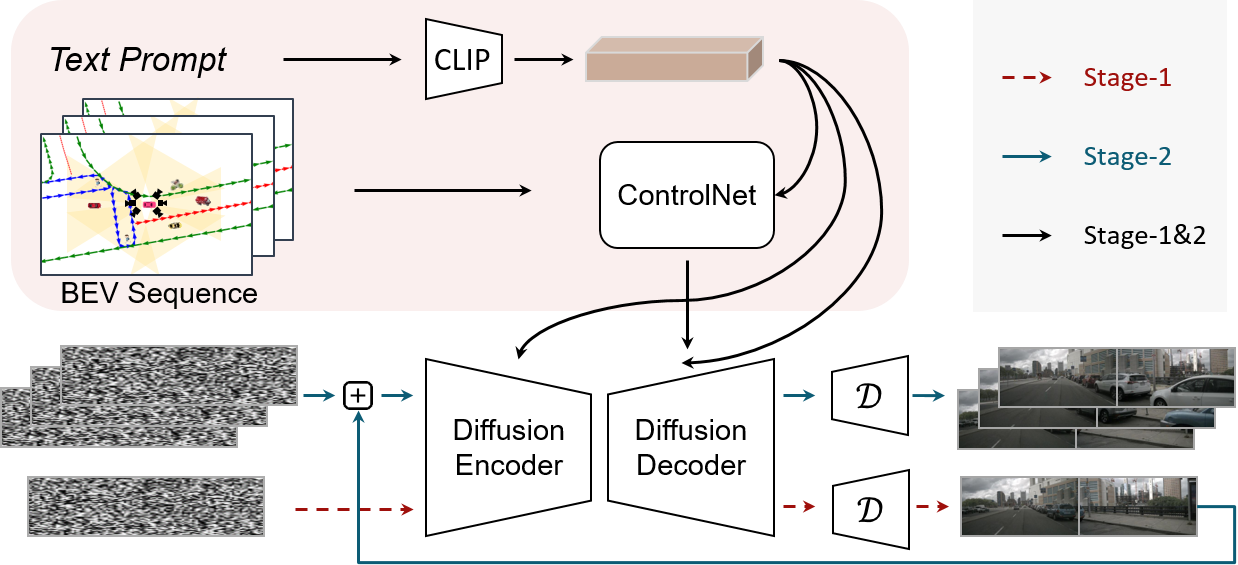}
   \hfill
    \caption{\textbf{The two-stage inference pipeline of Panacea.} Its two-stage process begins by creating multi-view images with BEV layouts, followed by using these images, along with subsequent BEV layouts, to facilitate the generation of following frames.
    }
    
   \label{fig:inference}
\end{figure}

\subsubsection{Decomposed 4D Attention}

The decomposed 4D attention-based UNet is designed to enable cross-view and cross-frame modeling while ensuring computational feasibility. A naive approach to multi-view video generation might employ 4D (HWVT) attention to exhaustively explore the multi-view video representations for coherent sample generation. However, this approach demands excessive memory and computational resources, surpassing the capabilities of even the most advanced A100 GPUs. Therefore, we propose a more efficient architecture called decomposed 4D attention, drawing inspiration from recent advancements in video representation learning~\cite{zhao2023td, DBLP:conf/icml/BertasiusWT21,DBLP:conf/iccv/Arnab0H0LS21,DBLP:conf/eccv/LinGZGMWDQL22}. Our model selectively retains the most critical attention operations: the attention between adjacent views and the attention among spatially aligned temporal patches. This leads to the introduction of two novel attention modules—cross-view attention and cross-frame attention—alongside the existing intra-view spatial attention. Empirical evaluations in Sec.~\ref{ablation} demonstrate that this decomposed 4D attention framework effectively generates coherent multi-view videos, maintaining both network feasibility and efficiency.

Fig.~\ref{fig:train} (b) details our decomposed 4D attention mechanism. The intra-view attention retains the design of the original spatial self-attention in the Stable Diffusion (SD) model, as formulated in Eq.~\ref{eq:intra}. To enhance cross-view consistency, we introduce cross-view attention. Our observations indicate that the correlation between adjacent views is paramount, while the correlation among non-adjacent views is comparatively less significant and can be disregarded. This cross-view attention is formulated in Eq.~\ref{eq:inter}. The cross-frame attention, mirroring the design of VLDM~\cite{DBLP:conf/cvpr/BlattmannRLD0FK23}, focuses on spatially aligned temporal patches. This component is crucial in endowing the model with temporal awareness, a key factor in generating temporally coherent videos.
\begin{equation}
\label{eq:intra}
\mathrm{Att}_{iv}\left(Q, K, V\right)=\operatorname{softmax}\left(\frac{Q_t^{v}\left(K_t^{v}\right)^{T}}{\sqrt{c}}\right) V_t^{v}
\end{equation}
\begin{equation}
\begin{aligned}
\label{eq:inter}
\mathrm {Att}_{cv}\left(Q, K, V\right) &= \operatorname{softmax}\left(\frac{Q_t^{v}\left([K_t^{v-1},K_t^{v+1}]\right)^{T}}{\sqrt{c}}\right) \\
&\quad \cdot [V_t^{v-1},V_t^{v+1}]
\end{aligned}
\end{equation}

Here $Q_t^{v},K_t^{v},V_t^{v}$ represent the queries, keys, and values within frame $t$ and view $v$, respectively.

\subsubsection{Two-Stage Pipeline}

To enhance the generation quality, we further adopt a two-stage training and inference pipeline. By bypassing the temporal-aware modules, our model could also operate as a multi-view image generator, which enables a unified architecture for two-stage video generation.

During training, we first train a separate set of weights dedicated to multi-view image generation. Then, as illustrated in Fig.~\ref{fig:train}, we train the second stage video generation weights, by concatenating a conditioned image alongside the diffused input. This conditioned image is integrated only with the first frame, while subsequent frames employ zero padding. Notably, in this second stage training, we employ ground truth images instead of the generated ones as condition. This approach equips our training process with an efficiency comparable to that of a single-stage video generation scheme.

During inference, as shown in Fig.~\ref{fig:inference}, we first sample multi-view frames using the weights of the first stage. This is followed by the generation of a multi-view video, which is conditioned on the initially generated frames, employing the weights of the second stage. This two-stage pipeline significantly enhances visual fidelity, a result attributable to the decomposition of spatial and temporal synthesis processes. The efficacy of this approach and its impact on visual quality will be further demonstrated in Sec.~\ref{ablation}.

\subsection{Generating Controllable Driving Scene Videos} \label{control}

In our Panacea model, designed for the advancement of autonomous driving systems, the controllability of synthesis samples emerge as a pivotal attribute. Panacea integrates two categories of control signals: a coarse-grained global control, encompassing textual attributes, and a fine-grained layout control, which involves BEV layout sequence. 

The coarse-grained global control endows the Panacea model to generate diverse multi-view videos.  This is achieved by integrating CLIP-encoded~\cite{DBLP:conf/icml/RadfordKHRGASAM21} text prompts into the UNet,  a method analogous to that used in Stable Diffusion. Benefiting from the Stable Diffusion pre-trained model, Panacea synthesizes specific driving scenes in response to textual prompts, as demonstrated in Fig.~\ref{fig:fig1} (c)

The Panacea model's fine-grained layout control facilitates the generation of synthesis samples that align with annotations. We use BEV layout sequences as the condition. Specifically, for a BEV sequence of duration T, we convert them into a perspective view and extract the control elements as object bounding boxes, object depth maps, road maps, and camera pose embeddings. Fig.~\ref{fig:train} (d) illustrates this process, where we employ different channels, represented by distinct colors, to delineate these segmented elements. This results in layout-controlling images with 19 channels: 10 for depth, 3 for bounding boxes, 3 for road maps, and 3 for camera pose embeddings. These 19-channel images are then integrated into the UNet using the ControlNet \cite{DBLP:journals/corr/abs-2302-05543} framework.

It is noteworthy that the camera pose essentially represents the direction vector~\cite{DBLP:conf/cvpr/ZhouK22,DBLP:conf/cvpr/00050GAG23,DBLP:journals/corr/abs-2301-04634}, which is derived from the camera's intrinsic and extrinsic parameters. Detailed construction of the camera pose is elaborated in the supplementary. This camera pose condition is incorporated to facilitate precise control over the viewpoint.

%% file: sections/4_experiments.tex
\section{Experiment}

\subsection{Datasets and Evaluation Metrics} \label{controllability}

We evaluate the generation quality and controllability of Panacea on the nuScenes~\cite{DBLP:conf/cvpr/CaesarBLVLXKPBB20} dataset. 

\noindent\textbf{nuScenes Dataset.} 
The nuScenes dataset, a public driving dataset, comprises 1000 scenes from Boston and Singapore.

Each scene is a 20 seconds video with about 40 frames. It offers 700 training scenes,  150 validation scenes, and 150 testing scenes with 6 camera views. The camera views overlap each other, covering the whole 360 field of view.

\noindent\textbf{Generation Quality Metrics.}
To evaluate the quality of our synthesised data, we utilize the frame-wise Fréchet Inception Distance (FID)~\cite{DBLP:conf/nips/HeuselRUNH17} and the Fréchet Video Distance (FVD)~\cite{DBLP:journals/corr/abs-1812-01717}, where FID reflects the image quality and FVD is a temporal-aware metric that reflects both the image quality and temporal consistency. 

We also use a matching-based~\cite{DBLP:conf/cvpr/SunSWBZ21} consistency score, View-Matching-Score (VMS), to measure the cross-view consistency of generated videos. This metric draws inspiration from a similar concept, the view-consistency-score (VCS), previously employed in BEVGen~\cite{DBLP:journals/corr/abs-2301-04634}. However, due to the unavailability of the original evaluation, we developed our own implementation of the VMS instead of directly using VCS.

\noindent\textbf{Controllability Metrics.} The controllability of Panacea is reflected by the alignment between the generated videos and the conditioned BEV sequences. To substantiate this alignment, we assess the perceptual performance on the nuScenes dataset, utilizing metrics such as the nuScenes Detection Score (NDS), mean Average Precision (mAP), mean Average Orientation Error (mAOE), and mean Average Velocity Error (mAVE). Our evaluation is two-fold: firstly, we compare the validation performance of our generated data against real data using a pre-trained perception model. Secondly, we explore the potential of augmenting training set as a strategy for performance enhancement.

We employ StreamPETR, a state-of-the-art (SoTA) video-based perception method, as our main evaluation tool. For the assessment of image-based generation approaches, we utilize StreamPETR-S, the single-frame variant of StreamPETR.

\begin{table}[tb]
    \centering
    \footnotesize
        \centering
        \resizebox{0.475\textwidth}{!}{
        \setlength{\tabcolsep}{8pt}
        \begin{tabular}{lcccc}
            \toprule
            Method     &Multi-View &Multi-Frame  & FVD$\downarrow$ & FID$\downarrow$ \\
            \hline
            BEVGen\cite{DBLP:journals/corr/abs-2301-04634} \  &$\checkmark$   &  &  & 25.54\\
            BEVControl\cite{DBLP:journals/corr/abs-2308-01661}  &$\checkmark$   &  & - & 24.85 \\
            DriveDreamer\cite{DBLP:journals/corr/abs-2309-09777}   & &$\checkmark$  & 452 & 52.6\\
            \hline
            \rowcolor[gray]{.9} 
            Panacea   &$\checkmark$  &$\checkmark$  & \textcolor{blue}{139}  & \textcolor{blue}{16.96}  \\
            \bottomrule
        \end{tabular}
        }
         \caption{Comparing FID and FVD metrics with SoTA methods on the validation set of the nuScenes dataset. }
        \label{tab:fvd}
\end{table}

\subsection{Implementation Details}

We implement our approach based on Stable Diffusion 2.1~\cite{DBLP:conf/cvpr/RombachBLEO22} . Pre-trained weights are used to initialize the spatial layers in UNet. During our two-stage training, the image weights of the first stage is optimized for 56k steps, and the video weights of the second stage is optimized for 40k steps. For inference, we utilize a DDIM~\cite{DBLP:conf/iclr/SongME21} sampler configured with 25 sampling steps. The video samples are generated at a spatial resolution of $256\times512$, with a frame length of 8. Correspondingly, our evaluation model, StreamPETR, building upon a ResNet50\cite{DBLP:conf/cvpr/HeZRS16} backbone, is trained at the same resolution of $256\times512$. More details can be found in supplementary material.

\subsection{Main Results}
\subsubsection{Quantitative Analysis}
\noindent\textbf{Generation Quality.} To verify the high fidelity of our generated results, we conduct a comparison of our approach  with various state-of-the-art driving scene generation methods. For fairness, we generate the whole validation set without using any post-processing strategies to select samples. Demonstrated in Tab.~\ref{tab:fvd}, our approach, Panacea, showcases remarkably superior generation quality, achieving an FVD of 139 and an FID of 16.96. These metrics significantly exceed those of all counterparts, encompassing both video-based method like DriveDreamer and image-based solutions such as BEVGen and BEVControl.

\noindent\textbf{Controllability for Autonomous Driving.} The controllability of our method is quantitatively assessed based on the perception performance metrics obtained using StreamPETR~\cite{DBLP:journals/corr/abs-2303-11926}.  We first generate the entire validation set of the nuSences by Panacea. Then, the perception performance is derived using a pre-trained StreamPETR model. The relative performance metrics, compared to the perception scores of real data, serve as indicators of the alignment between the generated samples and the conditioned BEV sequences. As depicted in Tab.~\ref{tab:valset}, Panacea achieves a relative performance of 68\%, underscoring a robust alignment of the generated samples. Furthermore, we present the first-stage results of our approach, which attains a relative performance of 72\%.

\begin{figure}
  \centering
   \includegraphics[width=0.9\linewidth]{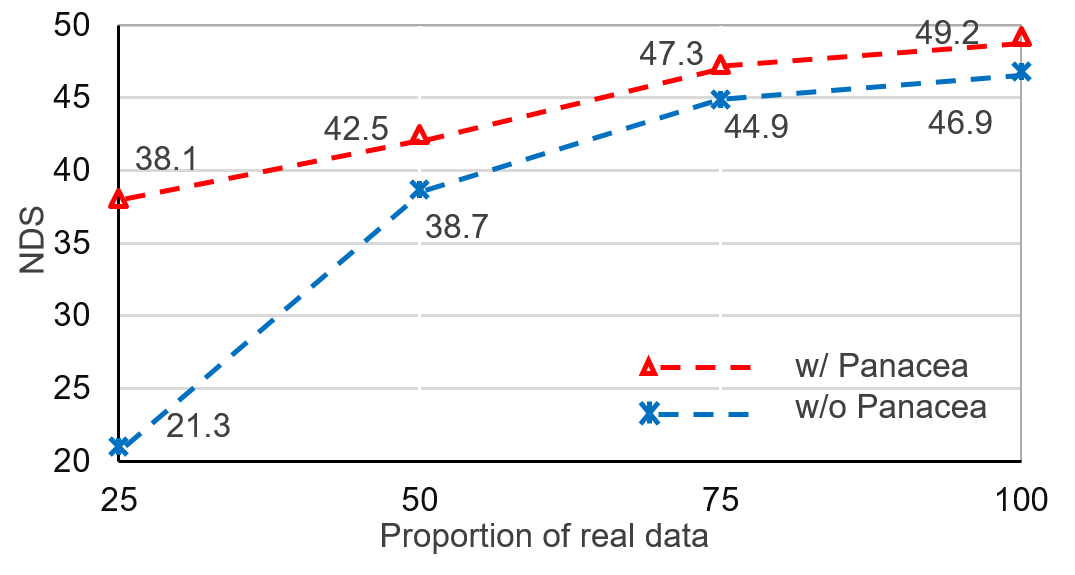}
   \hfill
   \caption{The detailed comparison of synthetic data augmentation across various real data ratios. We select portions of the real data at ratios of 25\%, 50\%, 75\%, and 100\%.}
   \label{proportion}
\end{figure}

Beyond the evaluation on the validation set, the more important feature of Panacea is its ability of generating an unlimited number of annotated training samples. Capitalizing on this, we synthesized a new training dataset for nuScenes, named Gen-nuScenes, to serve as an auxiliary training resource for the StreamPETR model. Intriguingly, the StreamPETR model trained exclusively on Gen-nuScenes achieved a notable nuScenes Detection Score (NDS) of 36.1\%, amounting to 77\% of the relative performance compared to the model trained on the actual nuScenes training set, as shown in Tab.~\ref{tab:augment}. More importantly, integrating generated data with real data propels the StreamPETR model to an NDS of 49.2, surpassing the model trained only on real data by 2.3 points. Additionally, Fig.~\ref{proportion} illustrates that Gen-nuScenes consistently bolsters the performance of StreamPETR across various real data ratios. These results collectively affirm that our Panacea model is adept at generating controllable multi-view video samples, constituting a valuable asset for autonomous driving systems.

We also demonstrate another application scenario of our Panacea model, which could elevate image-only datasets to video datasets by using their real images as condition. This elevation allows for the application of advanced video-based technologies. As indicated in Tab.~\ref{tab:expand}, this data elevation process yields a significant improvement, evidenced by a 5.8 point increase in NDS. Unfortunately, we also note a 2.3 point decrease in mAP, which we hypothesize is due to the inferior quality of the generated samples compared to the realistic ones and a domain mismatch between the generated training data and the real validation data. We hope that future advancements in improving the quality of generated samples will ameliorate this observed degradation.

\subsubsection{Qualitative Analysis}

\noindent\textbf{Temporal and View Consistency.}
As depicted in Fig.~\ref{fig:fig1} (a-b), Panacea shows the ability to generate the realistic multi-view videos directly from BEV sequences and text prompts. The generated videos exhibit notable temporal and cross-view consistency. For instance, as seen in Fig.~1 (a), the car in the front view maintains its appearance while approaching. Similarly, the content across different views is coherent, and the newly generated frames align seamlessly with the conditional frames (see Fig.~1 (b)). 

\begin{figure}
  \centering
   \includegraphics[width=\linewidth]{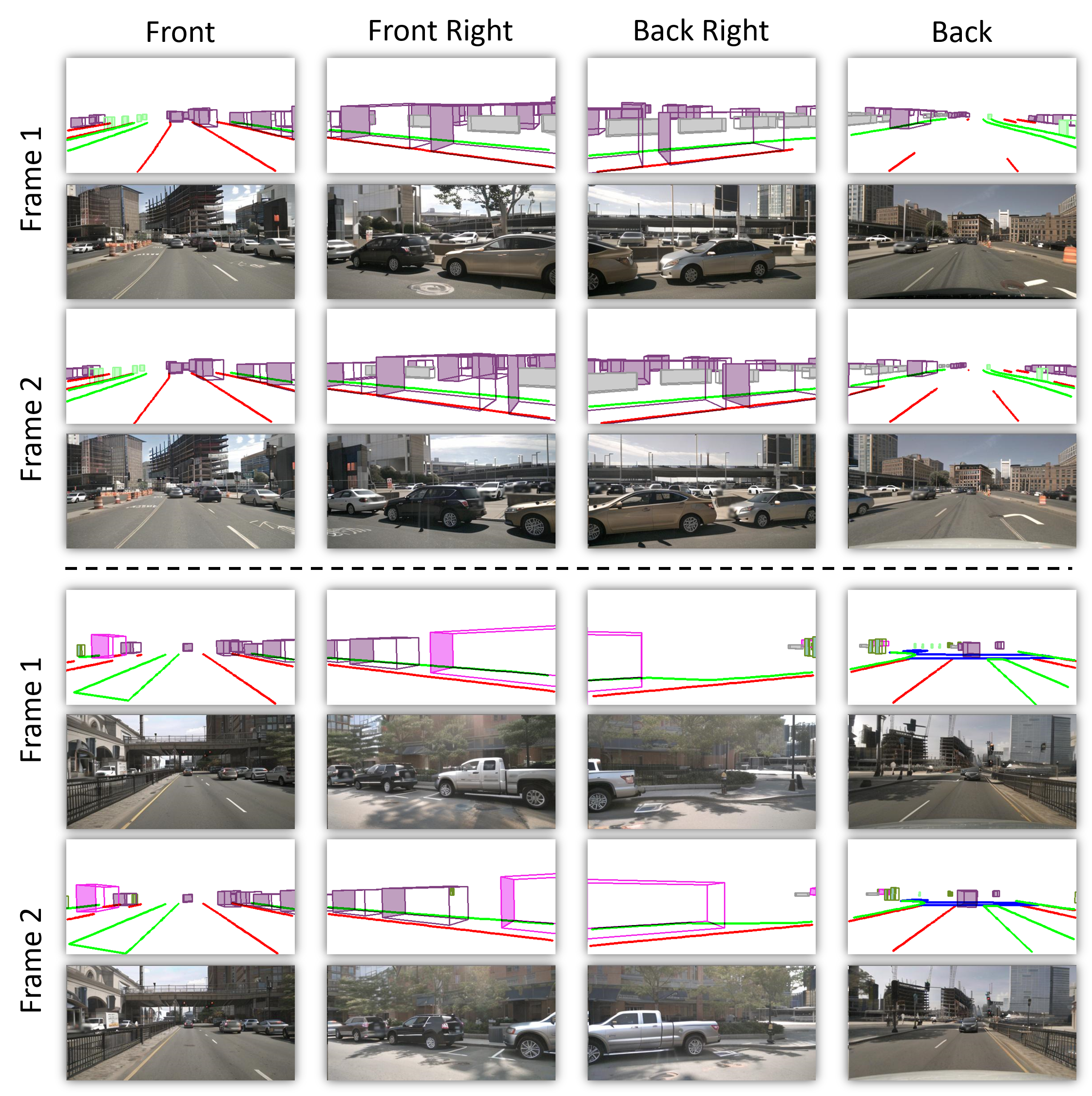}
   \hfill
   \caption{Controllable multi-view video generation. The column shows 4 different views and the rows shows adjacent frames aligned with BEV control.}
   \label{demo control}
\end{figure}

\noindent\textbf{Attribute and Layout Control.}
Fig.~\ref{fig:fig1} (c) illustrates the attribute control capabilities, where modifications to text prompts can manipulate elements like weather, time, and scene. This allows our approach to simulate a variety of rare driving scenarios, including extreme weather conditions such as rain and snow, thereby greatly enhancing the diversity of the data. 
Additionally, Fig.~\ref{demo control} depicts how cars and roads align precisely with the BEV layouts while maintaining excellent temporal and view consistency.

\begin{table}[tb]
    \centering
    \footnotesize
        \centering
        \resizebox{0.475\textwidth}{!}{
        \setlength{\tabcolsep}{7pt}
        \begin{tabular}{ccccc}
            \toprule
              Stage  &Image Size & Real & Generated &  NDS$\uparrow$\\
            \hline
             &512$\times$256  & $\checkmark$ & - & 34.3\\
           \cellcolor[gray]{.9} Panacea S1& \cellcolor[gray]{.9} 512$\times$256   & \cellcolor[gray]{.9} - &\cellcolor[gray]{.9}  $ \cellcolor[gray]{.9} \checkmark$ & \cellcolor[gray]{.9} \textcolor{blue}{24.7 (72\%)}\\
            \hline
             &512$\times$256   & $\checkmark$ & - &46.9 \\ 
               \cellcolor[gray]{.9} Panacea & \cellcolor[gray]{.9} 512$\times$256  &\cellcolor[gray]{.9} - &\cellcolor[gray]{.9}  $ \cellcolor[gray]{.9} \checkmark$  & \cellcolor[gray]{.9} \textcolor{blue}{32.1 (68\%) }\\
            \bottomrule
        \end{tabular}
        }
         \caption{Comparison of the generated data with real data on the validation set, employing a pre-trained perception model. The first stage (S1) generates the single-frame image, while the second stage (S2) outputs the multi-frame video. The evaluation of image is carried out by StreamPETR-S (single frame), whereas the video data is assessed through StreamPETR.}
        \label{tab:valset}
\end{table}

\begin{table}[tb]
    \centering
    \footnotesize
        \centering
        \resizebox{0.475\textwidth}{!}{
        \setlength{\tabcolsep}{4pt}
        \begin{tabular}{cccccc}
            \toprule
            Real &  Generated      & mAP$\uparrow$ & mAOE$\downarrow$ & mAVE$\downarrow$ & NDS$\uparrow$ \\
            \hline
            $\checkmark$  & - & 34.5	& 59.4 & 29.1  & 46.9 \\
             -  &$\checkmark$  &22.5 &72.7 & 46.9 & 36.1 \\
            \hline
            \rowcolor[gray]{.9} 
            $\checkmark$  & $\checkmark$   & 37.1 \textcolor{blue}{(+2.6\%)} &54.2 &27.3 &49.2 \textcolor{blue}{(+2.3\%)} \\
            \bottomrule
        \end{tabular}
        }
         \caption{Comparison involving data augmentation using synthetic data. We attempt training exclusively using synthetic data and also explore integrating it with real data. }
        \label{tab:augment}
\end{table}

\begin{table}[tb]
    \centering
    \footnotesize
        \centering
        \resizebox{0.475\textwidth}{!}{
        \setlength{\tabcolsep}{2.0pt}
        \begin{tabular}{ccccccc}
            \toprule
            Multi-Frame & Real &  Generated & mAP$\uparrow$ & mAOE$\downarrow$ & mAVE$\downarrow$ & NDS$\uparrow$ \\
            \hline
            - & $\checkmark$  & - & 28.6	& 67.9 & 101.6  & 34.3 \\
            \rowcolor[gray]{.9} 
           $\checkmark$  & -  &$\checkmark$  &26.3 \textcolor{blue}{(-2.3\%)} &61.1 &42.9 &40.1 \textcolor{blue}{(+5.8\%)} \\
            \bottomrule
        \end{tabular}
        }
         \caption{Effect of elevating image-only data into video dataset.}
        \label{tab:expand}
\end{table}

\subsection{Ablation Studies} \label{ablation}
This section validates two important designs in Panacea: decomposed 4D attention and the two-stage pipeline.

\begin{figure}
  \centering
   \includegraphics[width=\linewidth]{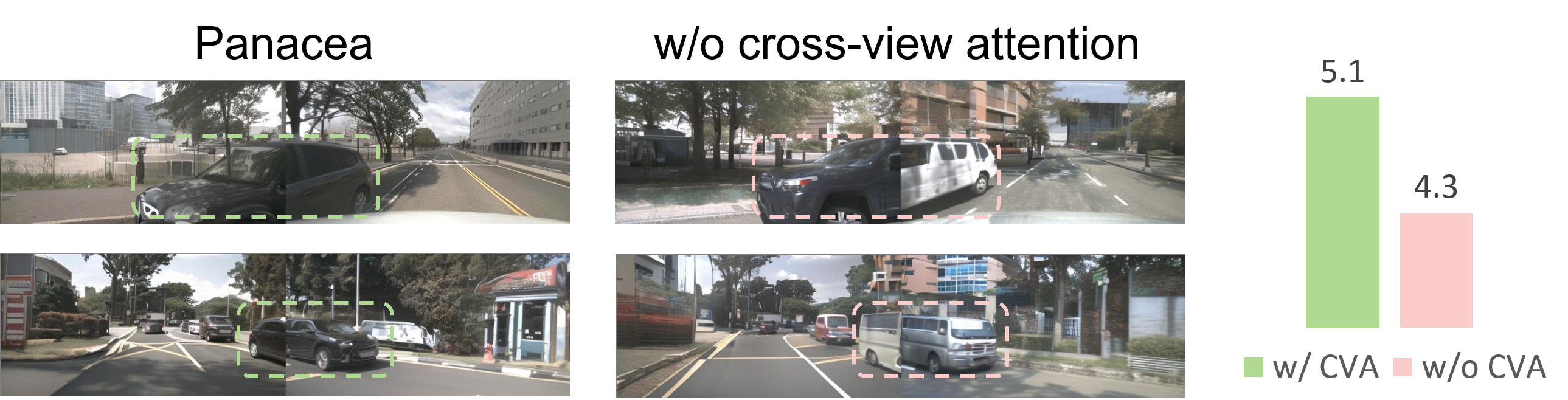}
   \hfill
   \caption{Ablation on the efficiency of cross-view attention (CVA). The cars in the generated frames with or without cross-view attention are highlighted by green and pink dashed boxes, respectively. }
   \label{cross view}
\end{figure}

\noindent \textbf{Decomposed 4D Attention.}
We first investigate the impact of the cross-view attention mechanism in decomposed 4D attention. As illustrated in Tab.~\ref{tab:ablation}, the exclusion of the cross-view module results in a degradation of 108 and 5.11 in FVD and FID, respectively. This indicates the crucial role of the cross-view module in improving video quality. Additionally, to assess view consistency more precisely, we evaluate the VMS metric. Incorporating the cross-view module results in a 0.8 point enhancement, corroborating its efficacy in improving multi-view consistency, as showcased in the right of Fig.~\ref{cross view}.

The effectiveness of the cross-view module is also illustrated in the left of Fig.~\ref{cross view}. Without the cross-view module, the appearance of cars are inconsistent across different views. Conversely, with the integration of the cross-view module, there is a significant improvement in maintaining the consistency of cars and scenes across views.

\begin{table}[tb]
    \centering
    \footnotesize
        \centering
        \resizebox{0.475\textwidth}{!}{
        \setlength{\tabcolsep}{23pt}
        \begin{tabular}{lcc}
            \toprule
            Settings        & FVD$\downarrow$ & FID$\downarrow$  \\
            \hline
            \rowcolor[gray]{.9} 
            Panacea   & 139 & 16.96 \\
            \hline
            w/o Cross-view    & 247 \textcolor{olive}{(+108)} & 22.07 \textcolor{olive}{(+5.11)}\\ 
            w/o Cross-frame   & 214 \textcolor{olive}{(+75)}& 20.43 \textcolor{olive}{(+3.47)}\\
            w/o Two stage & 305 \textcolor{olive}{(+166)}& 36.61 \textcolor{olive}{(+19.65)}\\           
            \bottomrule
        \end{tabular}
        }
        \caption{Ablation studies on different settings. 'w/o Two stage' denotes the method where a video generation model is directly trained and inferred without an initial stage of image generation.}

        \label{tab:ablation}
\end{table}

\begin{figure}
  \centering
   \includegraphics[width=1.0\linewidth]{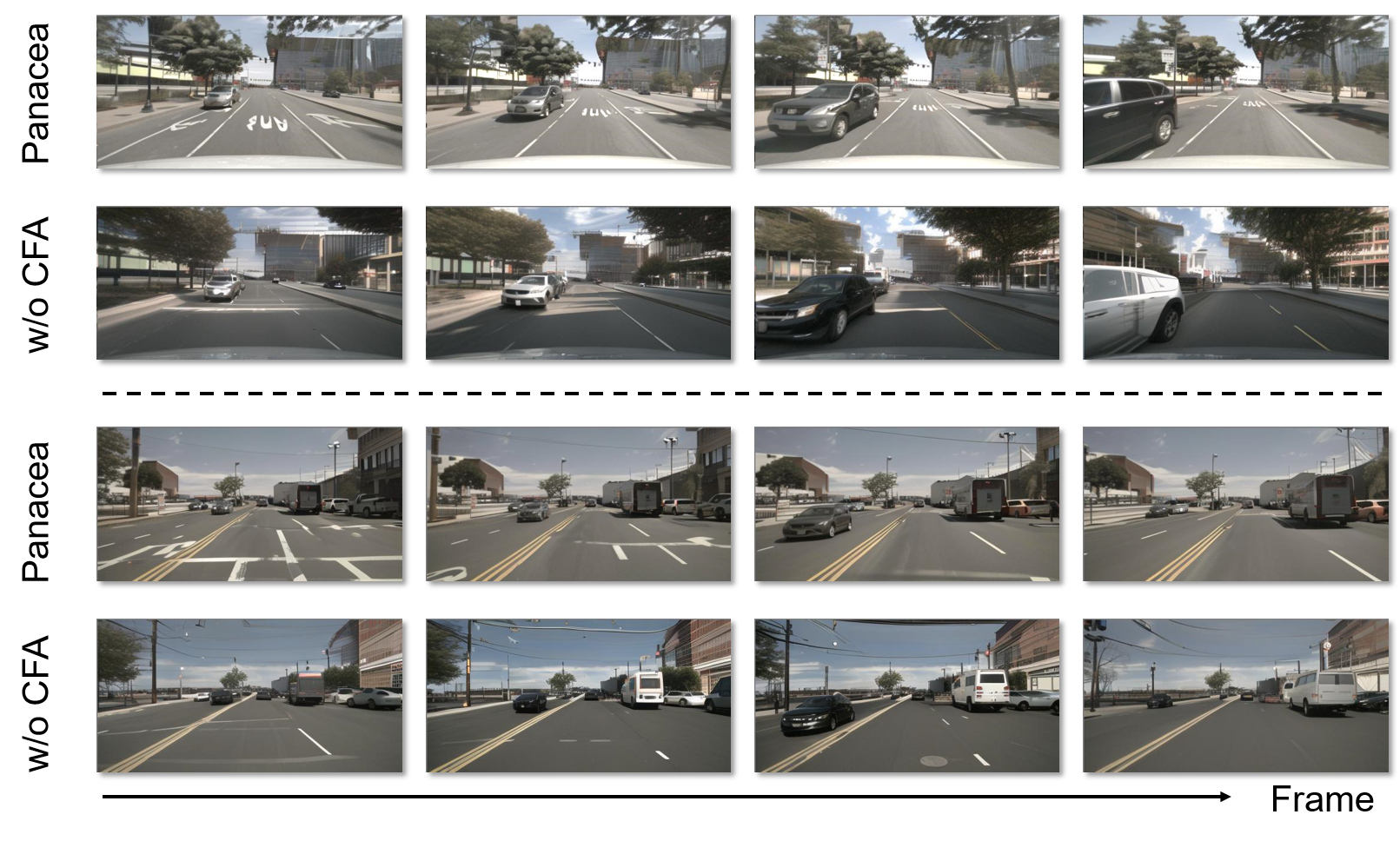}
   \hfill
   \caption{Ablation study on the effects of Cross-Frame Attention (CFA). The sequential frames are displayed from left to right.}
   \label{temporal}
   \vspace{-0.5cm}
\end{figure}

To evaluate the temporal attention, we perform an ablation study by eliminating this component. As shown in Tab.~\ref{tab:ablation}, a notable degradation of 75 points in FVD is observed when the temporal module is removed, highlighting its crucial role in maintaining temporal consistency. Furthermore, Fig.~\ref{temporal} concisely demonstrates that without the cross-frame attention module, the model fails to retain temporal consistency, as evident in the variations of the car's appearance across frames.

\noindent \textbf{Two Stage Pipeline.}
To confirm that our two-stage pipeline enhances generation quality, we conduct a comparison against the single-stage schema. The single-stage approach yields an FID of 36.61 and an FVD of 305,  markedly inferior to the results achieved by Panacea. 

This comparison unequivocally underscores the critical impact of the two-stage pipeline in substantially elevating the quality of the generated videos.

%% file: sections/5_conclusion.tex
\section{Conclusion}
We propose Panacea, a cutting-edge generator meticulously designed to create manipulable panoramic videos for driving scenarios. Within this innovative framework, we incorporate a decomposed 4D attention module to ensure temporal and cross-view consistency, facilitating the generation of realistic multi-view images and videos. Besides, a two-stage training strategy is employed to further enhance the generation quality. Significantly, Panacea is adept at handling a variety of control signals to produce videos with precise annotations. Through extensive experiments, Panacea has demonstrated its proficiency in generating high-quality, well-annotated panoramic driving-scene videos. These videos are invaluable, serving not only in BEV perception but also hold promise in real-world driving simulations. Looking ahead, we aspire to delve into the expansive potential of Panacea in real-world simulation, and integrate control signals with more diversity.

%% file: sup/sup.tex
\section{More Implementation Details}

\begin{table*}[tb]
    \centering
    \footnotesize
        \centering
        \resizebox{0.95\textwidth}{!}{
        \setlength{\tabcolsep}{3pt}
        \begin{tabular}{c|c|ccccccccccccc}
            \toprule
              &mAP  &car & truck & bus & trailer &constr. &pedestrian & motorcycle & bicycle & cone & barrier \\
            \hline
             Real  &34.5 &54.5 & 28.9 & 30.1 & 8.4 &9.7 &39.9 & 34.4 & 52.7 & 56.6 & 50.1 \\
            \hline
            \rowcolor[gray]{.9} 
             +Panacea  &37.1\textcolor{blue}{ +2.6} &57.1\textcolor{blue}{ +2.6} & 29.4\textcolor{blue}{ +0.5} &30.7\textcolor{blue}{ +0.6} & 14.6\textcolor{blue}{ +6.2} &10.8\textcolor{blue}{ +1.1} &42.7\textcolor{blue}{ +2.8} & 38.7\textcolor{blue}{ +3.3} & 50.4\textcolor{blue}{ -2.3} & 60.8\textcolor{blue}{ +4.2} & 54.7\textcolor{blue}{ +4.6} \\
            \bottomrule
        \end{tabular}
        }
         \caption{Per-class comparison involving data augmentation using synthetic
data, where 'constr' is short for construction vehicles.}
        \label{tab:valset1}
\end{table*}

\begin{figure}
\vspace{0.7cm}
  \centering
   \includegraphics[width=1.0\linewidth]{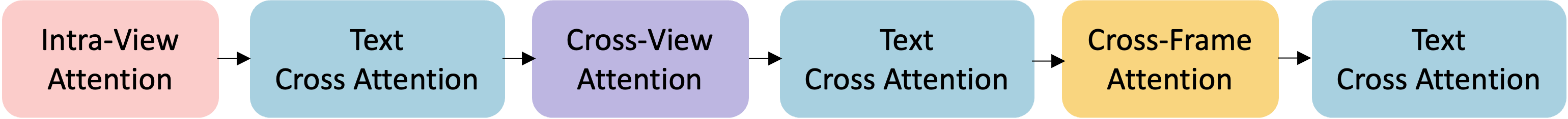}
   \hfill
   \caption{Illustration of the structure of decomposed 4D attention}
   \label{4d}
\end{figure}

\noindent\textbf{Decomposed 4D Attention.}
We present a more detailed demonstration of our decomposed 4D attention, as shown in Fig.~\ref{4d}, we incorporate the text context into the intra-view, inter-view, and cross-frame blocks by utilizing cross attention. This approach is consistent with the methodology employed in \cite{DBLP:conf/cvpr/RombachBLEO22}.

\noindent\textbf{Construction of Camera Pose.}
In order to facilitate consistency across multi-view images, , we introduce a camera pose prior into control signals by encoding the camera pose into the 3D direction vector following the success in 3D perception~\cite{zhou2022cross, liu2022petr}. Specifically, for each point $p^{c} = (u,v,d, 1)$ in camera frustum space, it can be projected into 3D space by intrinsic $K\in \mathbb{R}^{4 \times 4}$ and extrinsic $E\in \mathbb{R}^{4 \times 4}$ as in Eq.~\ref{eq_proj}:

\begin{equation}\label{eq_proj}
p^{3d} = E \times K^{-1} \times p^{c}
\end{equation}

\begin{figure}
  \centering
   \includegraphics[width=0.9\linewidth]{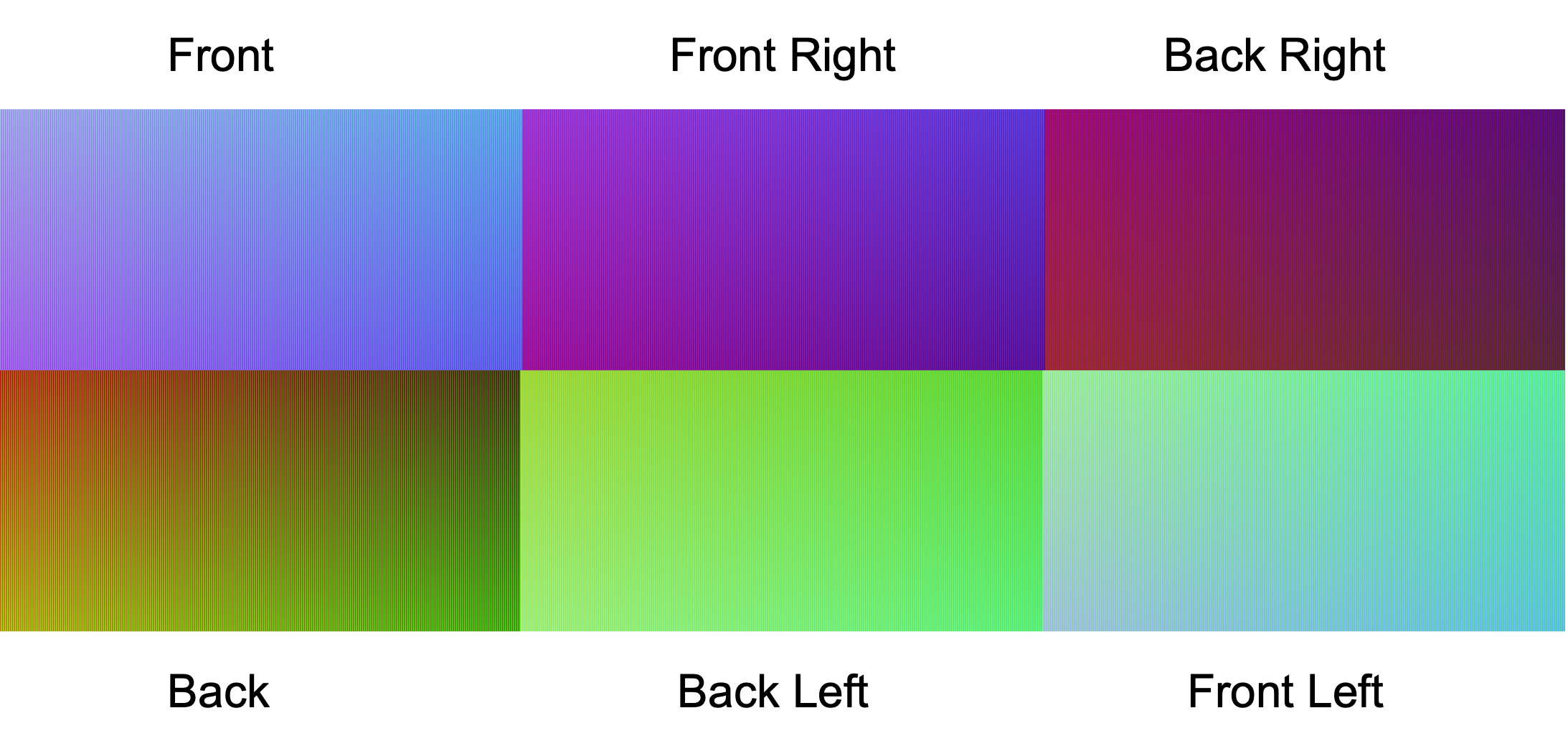}
   \hfill
   \caption{Visualization of pseudo-color image of camera pose.}
   \label{camera}
\end{figure}

Here $(u, v)$ denotes the pixel coordinates in the image, $d$ is the depth along the axis orthogonal to the image plane, $1$ is for the convenience of calculation in homogeneous form. We pick two points and set the depth of them to $d_1=1, d_2 =2$. Then the direction vector of the corresponding camera ray can be computed by:
\begin{equation}
\text{DV}(u,v) = p^{3d} (d_2) - p^{3d} (d_1)
\end{equation}
We normalize the direction vector by dividing the vector magnitude and multiply $255$ to simply convert it into RGB pseudo-color image, as shown in Fig.~\ref{camera}.

\noindent\textbf{Training Details of StreamPETR.}
StreamPETR is trained at a resolution of  512$\times$256 instead of the 704$\times$256 in  original baseline \cite{DBLP:journals/corr/abs-2303-11926}. For the single-frame baseline (StreamPETR-S), we close the query propagation and retrain the model. The batch size is 16 and the learning rate is 4e-4. 
\section{More Experimental Results}
\noindent\textbf{BEV Sequence Control.}
Given that ControlNet~\cite{DBLP:journals/corr/abs-2302-05543} potentially imposes an additional cost, we explore a simpler way to introduce BEV control signals without ControlNet, merely through the concatenation of the BEV feature with the input latent codes in the channel dimension. As show in Tab.~\ref{tab:control}, the FVD and FID are 87 and 3.19 higher than using ControlNet, thus we keep the ControlNet as our default design to introduce BEV control.

\begin{table}[tb]
    \centering
    \footnotesize
        \centering
        \resizebox{0.475\textwidth}{!}{
        \setlength{\tabcolsep}{23pt}
        \begin{tabular}{lcc}
            \toprule
            Settings        & FVD$\downarrow$ & FID$\downarrow$  \\
            \hline
            \rowcolor[gray]{.9} 
            Panacea (ControlNet)   & 139 & 16.96 \\
            Panacea (Concat) &226 \textcolor{olive}{(+87)} & 20.15 \textcolor{olive}{(+3.19)}\\      
            \bottomrule
        \end{tabular}
        }
        \caption{Ablation studies on how to introduce BEV control signals.}
    
        \label{tab:control}
\end{table}

\noindent\textbf{Per-class Results.}
We provide per-class detection results in Tab.~\ref{tab:valset1} on the validation set of nuScenes with StreamPETR, corresponding to the results in Tab. 3 in the main paper. As shown in Tab.~\ref{tab:valset1}, incorporating data generated by Panacea leads to enhanced AP for virtually all classes.

\section{More Visualization Results}
\begin{figure*}
  \centering
   \includegraphics[width=0.9\linewidth]{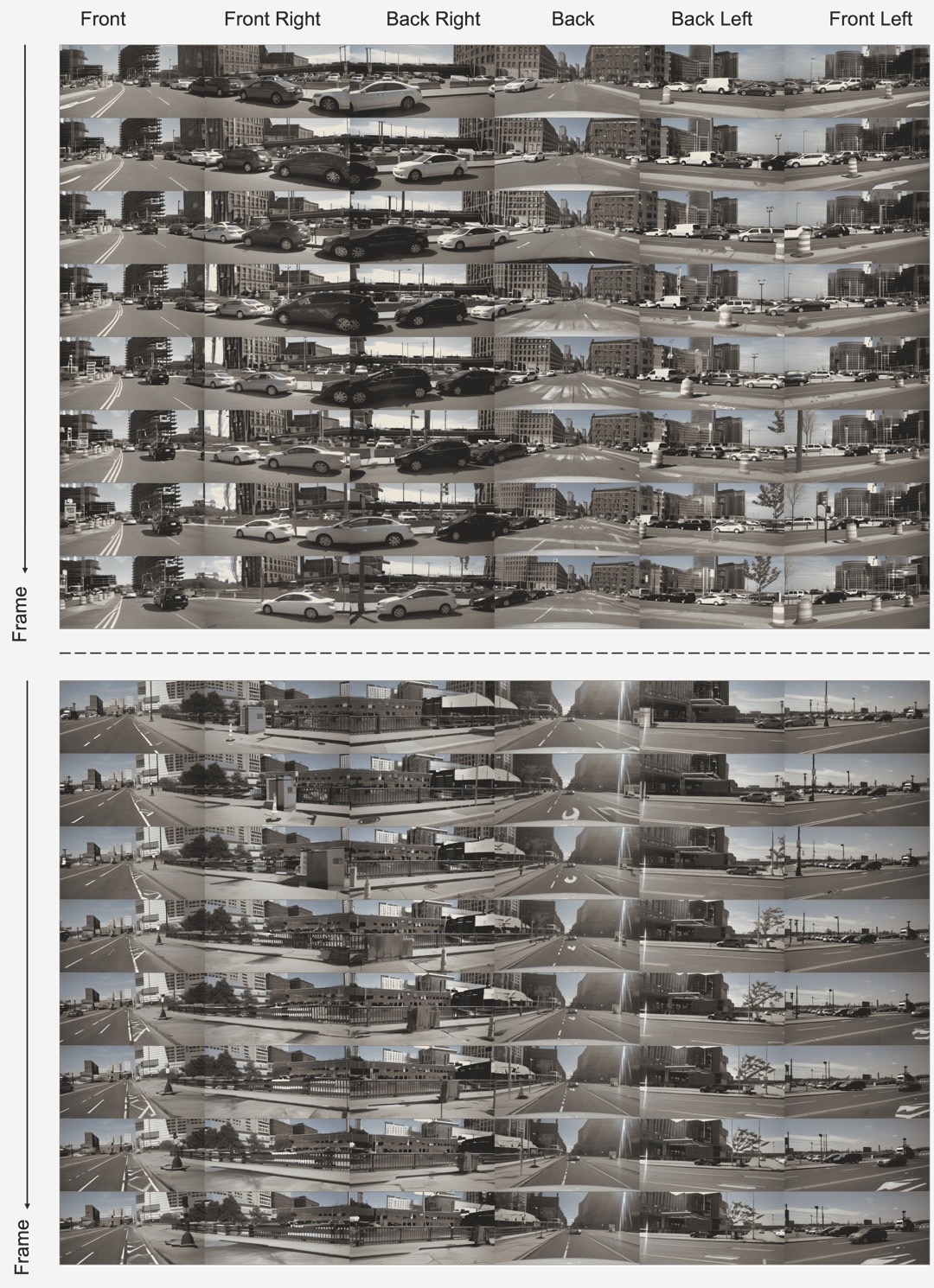}
   \hfill
   \caption{Multi-view videos generated by Panacea.}
   \label{demo48}
\end{figure*}

\noindent\textbf{Multi-View Video Generation.}
We present more visualization results of our Panacea. As shown in Fig.~\ref{demo48}, we exhibit multi-view videos generated from the validation set, comprising a total of 8 frames and 6 views per video. One can observe that these generated videos maintain good temporal and cross-view consistency.

\begin{figure*}
  \centering
   \includegraphics[width=0.95\linewidth]{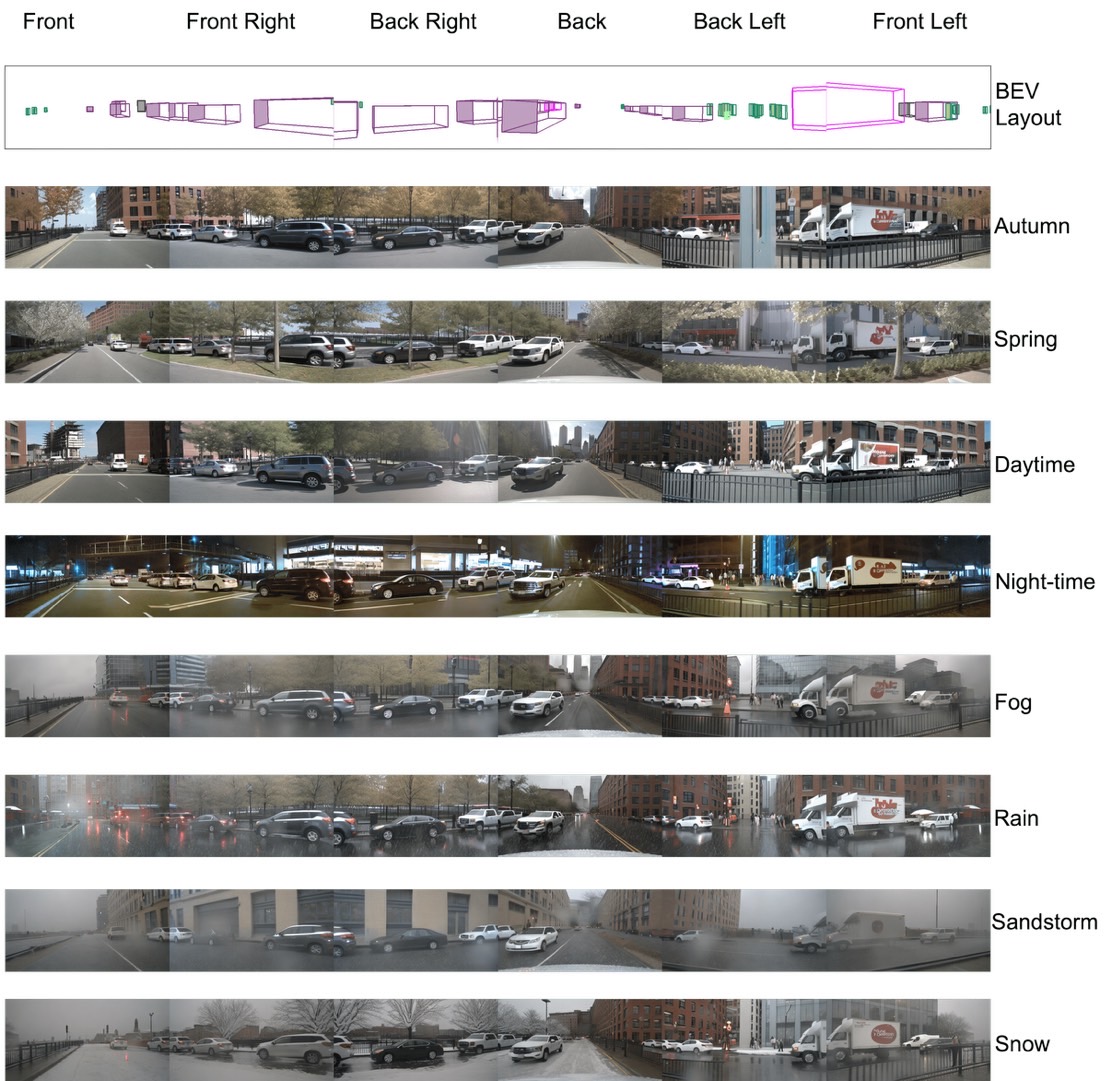}
   \hfill
   \caption{Multi-view frames generated by Panacea with different attribute controls.}
   \label{weather}
\end{figure*}

\begin{figure*}
  \centering
   \includegraphics[width=1.0\linewidth]{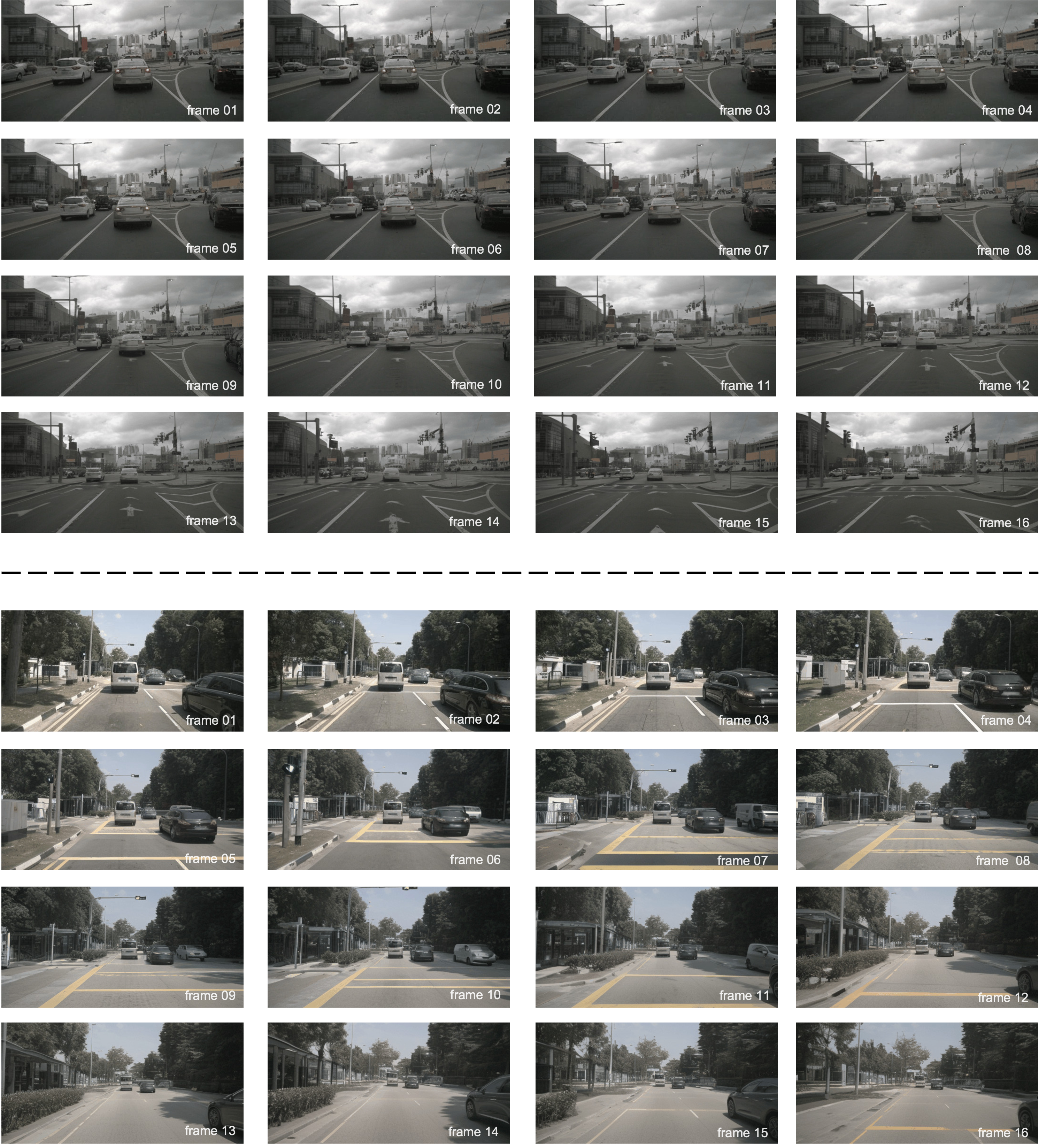}
   \hfill
   \caption{Long videos generated by Panacea, from top to bottom and left to right are consecutive frames.}
   \label{long}
\end{figure*}

\noindent\textbf{Attribute Controllable Video Generation.}
We present visualization results with different attributes control. As illustrated in Fig.~\ref{weather}, we exhibit the same case with varying attribute control. Our model is capable of generating multi-view frames that align well with BEV layout and global attributes.

\noindent\textbf{Generating Long Videos.}
Our Panacea possesses the potential to serve as a viable candidate for real-world simulation. We conduct an experiment in which we discard the BEV layout control part and use four frames as image conditions, applying iterative inference with a sliding window to generate long videos. As shown in Fig.~\ref{long}, we exhibit consecutive frames of the generated videos. Panacea is proficient in generating high-quality long videos.

\section{More Related Works}
There are two main categories of driving scene synthesis methods, one is the generative methods based on GAN or diffusion models~\cite{DBLP:journals/corr/abs-2308-01661,DBLP:journals/corr/abs-2301-04634}, as expounded in the main paper, and the other is NeRF (Neural Radiance Fields) based methods, such as Unisim~\cite{yang2023unisim}, Neuralsim~\cite{heiden2021neuralsim} and Mars~\cite{wu2023mars}. The generation quality of NeRF-based methods can be high. However, the most significant difference between this and our diffusion-based method is that NeRF-based methods can merely reconstruct pre-existing scenes that the model is trained on, consequently resulting in a lack of diversity.

\section{Gen-nuScenes Dataset}
We synthesis a new training dataset named Gen-nuScenes to enhance the training of Stream-PETR, which contains totally 139440 videos, each video of 8 frame length.

\section{Limitations}
Our task addresses a novel problem of controllable multi-view video generation for autonomous driving. Due to time and resource constraints, we have not yet undertaken more detailed designs of the model. Consequently, the quality of the videos generated by our model still leaves room for improvement. For example, the temporal and view consistency are not perfect since learning the correlation between all views and all frames poses a significant challenge, especially when the frame length is long. In the future, more effective designs for temporal and view consistency might need to be explored. Additionally, Panacea's computational cost of inference is relatively high. Thus we plan to enhance the efficiency of Panacea in the future. Furthermore, we still employ a relatively low spatial resolution due to time and resource limitations. Future work could involve integrating more powerful generative models, such as SD-XL \cite{DBLP:journals/corr/abs-2307-01952} and more efficient manners to produce high-fidelity videos of larger spatial resolution.